\documentclass{article}

\usepackage[final,nonatbib]{mlsafety_neurips_2022}

\usepackage[utf8]{inputenc} % allow utf-8 input
\usepackage[T1]{fontenc}    % use 8-bit T1 fonts
\usepackage{hyperref}       % hyperlinks
\usepackage{url}            % simple URL typesetting
\usepackage{booktabs}       % professional-quality tables
\usepackage{amsfonts}       % blackboard math symbols
\usepackage{nicefrac}       % compact symbols for 1/2, etc.
\usepackage{microtype}      % microtypography
\usepackage{xcolor}         % colors
\usepackage{graphicx}       
\usepackage{amsmath} 

\newcommand\figref{Figure~\ref}
\newcommand\secref{Section~\ref}
\newcommand\apref{Appendix~\ref}

\title{Diagnostics for Deep Neural Networks with\\Automated Copy/Paste Attacks}

\author{
  Stephen Casper,$^{*}$ Kaivalya Hariharan$^{*}$ Dylan Hadfield-Menell\\
  MIT CSAIL\\
  \texttt{\{scasper, kaivu, dylanhm\} @mit.edu}\\
  $*$ Equal Contribution
}

\begin{document}

\maketitle

\begin{abstract}

This paper considers the problem of helping humans exercise scalable oversight over deep neural networks (DNNs). Adversarial examples can be useful by helping to reveal weaknesses in DNNs, but they can be difficult to interpret or draw actionable conclusions from. Some previous works have proposed using human-interpretable adversarial attacks including \emph{copy/paste} attacks in which one natural image pasted into another causes an unexpected misclassification. We build on these with two contributions. First, we introduce Search for Natural Adversarial Features Using Embeddings (SNAFUE) which offers a fully automated method for finding copy/paste attacks. Second, we use SNAFUE to red team an ImageNet classifier. We reproduce copy/paste attacks from previous works and find hundreds of other easily-describable vulnerabilities, all without a human in the loop. \footnote{\href{https://github.com/thestephencasper/snafue}{https://github.com/thestephencasper/snafue}}

\end{abstract}

\section{Introduction} \label{sec:introduction}

It is important to have scalable methods that allow humans to exercise effective oversight over deep neural networks. 
Adversarial examples are one type of tool which can be used to study weaknesses in DNNs \cite{dong2017towards, tomsett2018failure, raukur2022toward}.
Typically, adversarial examples are generated by optimizing perturbations to the input of a network.
And some previous works offer examples of adversaries being used to develop generalizable interpretations of DNNs \cite{dong2017towards, tomsett2018failure, ilyas2019adversarial, casper2022robust}.

However, there are limitations to what one can learn about flaws in DNNs from synthesized features \cite{borowski2020exemplary}. 
First, synthetic adversarial perturbations are often difficult to describe and thus offer limited help with human-centered approaches to interpretability. 
Second, even when synthetic adversarial features are interpretable, it is unclear without additional testing whether they fool a DNN due to their interpretable features or due to hidden motifs  \cite{brown2017adversarial, ilyas2019adversarial}.
This makes developing a generalizable understanding from them difficult. 
Third, there is a gap between research and practice in adversarial robustness \cite{apruzzese2022real}.
Real-world failures of DNNs are often due to atypical natural features or combinations thereof \cite{hendrycks2021natural}, but synthesized features are off this distribution.

\begin{figure*}[t!]
    \centering
    \includegraphics[width=0.95\linewidth]{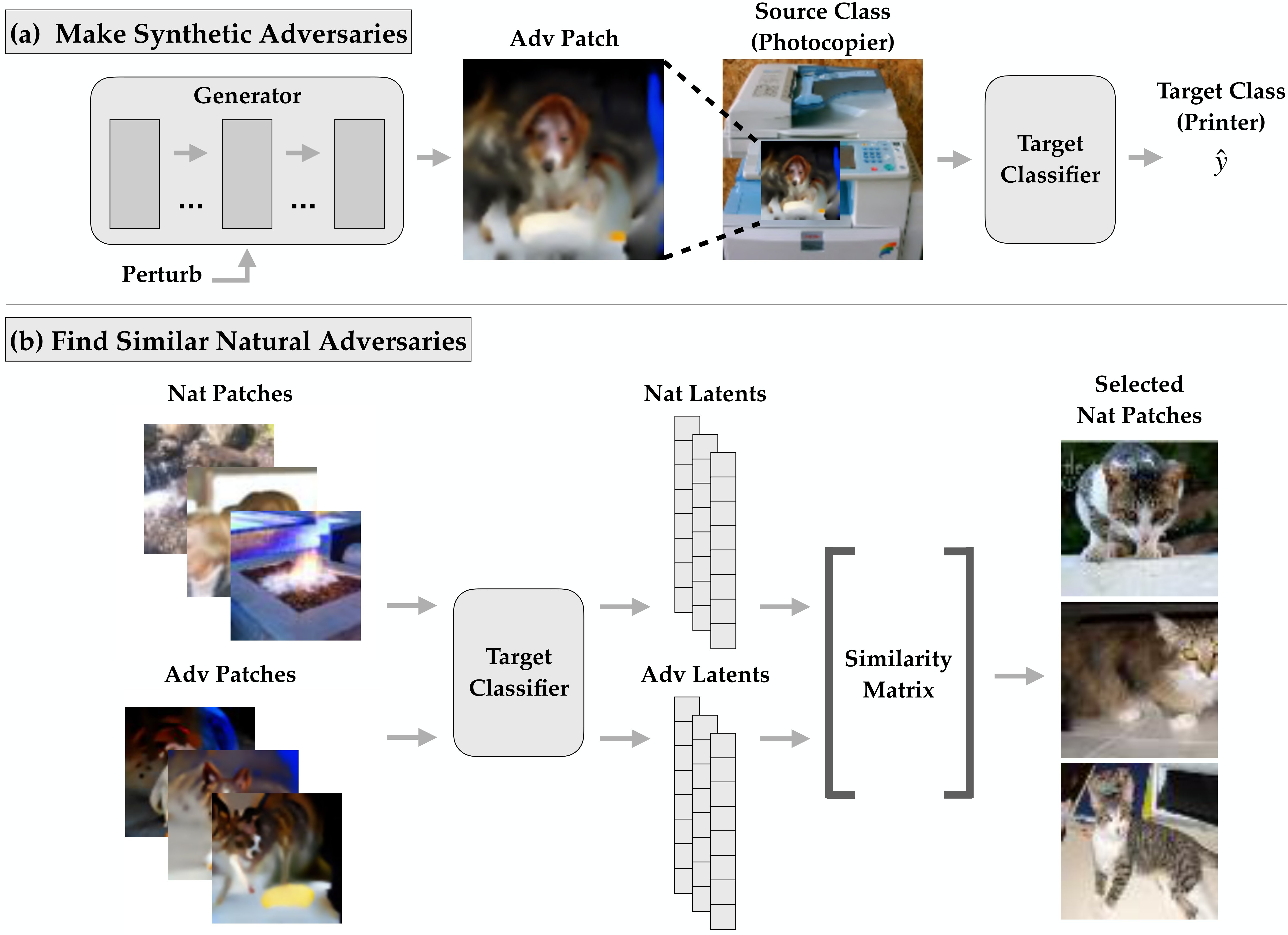}
    \caption{SNAFUE, our automated method for finding targeted copy/paste attacks. This example illustrates an experiment which found that cats can make photocopiers misclassified as printers. (a) First, we create feature level adversarial patches as in \cite{casper2022robust} by perturbing the latent activations of a generator. (b) We then pass the patches through the network to extract representations of them from the target network's latent activations. Finally, we select the natural patches whose latents are the most similar to the adversarial ones.}
    \label{fig:fig1}
\end{figure*}

\begin{figure*}
    \centering
    \includegraphics[width=0.99\linewidth]{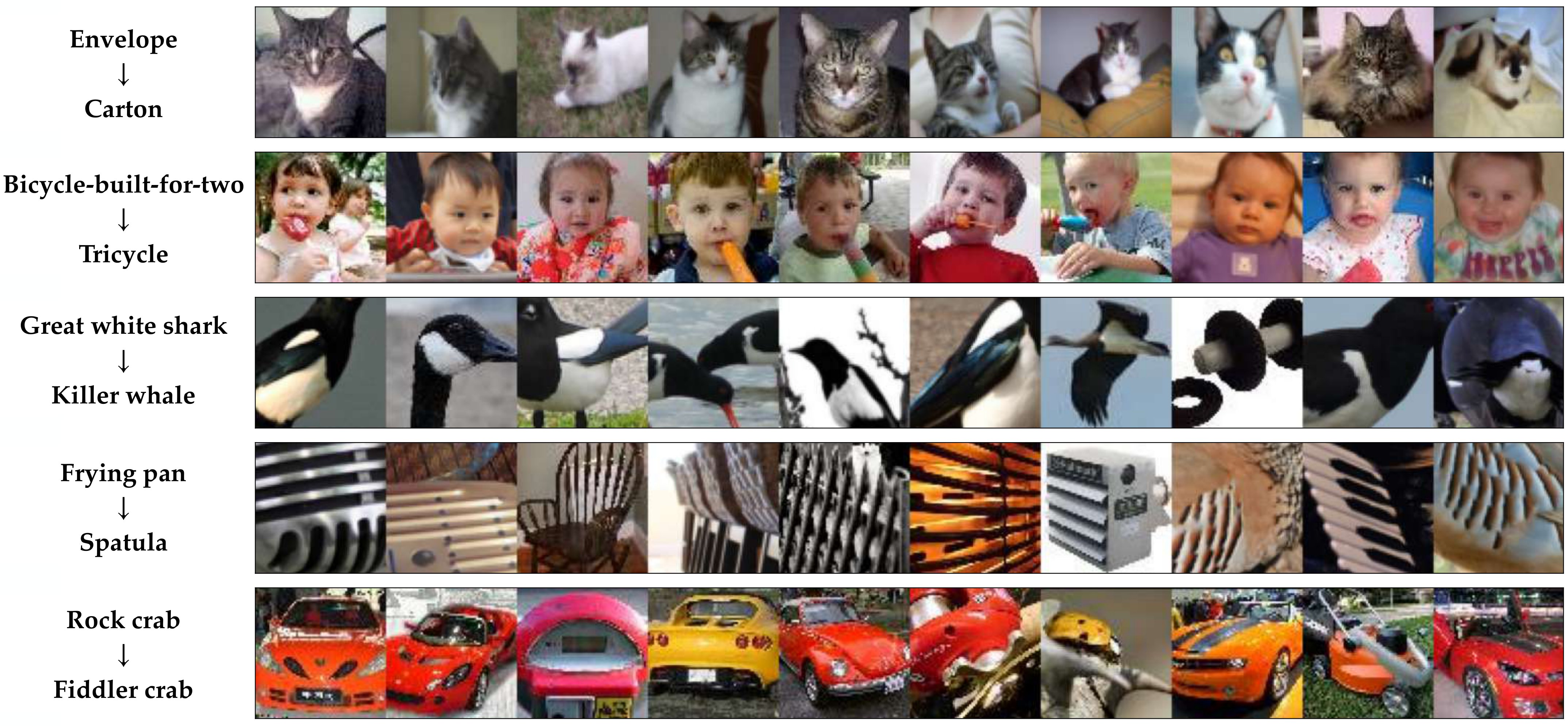}
    \caption{Examples of targeted natural adversarial patches for image classification identified using SNAFUE. They reveal consistent, easily-describable failure modes that can be used to interpret the network (e.g. ``envelopes plus cats are misclassified by the network as cartons''). Each row contains 10 patches labeled with the attack source and target. When a patch is inserted into any source class image, it tends to cause misclassification as the target class. See \figref{fig:breadth} for quantitative evaluation and \figref{fig:nat_examples2} in the Appendix for additional examples.}
    \label{fig:nat_examples1}
\end{figure*}

Here, we work to diagnose weaknesses in DNNs using \emph{natural}, \emph{interpretable} features.
We introduce using a Search for Natural Adversarial Features Using Embeddings (SNAFUE) to find novel adversarial combinations of natural features. 
We apply SNAFUE to find \emph{copy/paste} attacks for an image classifier in which one natural image is inserted as a patch into another to induce a targeted misclassification. 
\figref{fig:fig1} outlines this approach. 
First, we use a generator to synthesize robust feature-level adversarial patches \cite{casper2022robust} which are designed to make any image from a particular source class misclassified as a target. 
Second, we use the target model's latent activations to create embeddings of both these synthetic patches and a dataset of natural patches.
Finally we select the natural patches that embed most similarly to the synthetic ones.

We apply SNAFUE at the ImageNet scale.
First, we use SNAFUE to replicate all successful known examples of copy/paste attacks from previous works with no human involvement.
Second, we demonstrate its scalability by identifying hundreds of vulnerabilities. 
\figref{fig:nat_examples1} and \figref{fig:nat_examples2} show examples which illustrate easily-describable misassociations between features and classes in the network. 
Overall, this work makes two contributions. 
\begin{enumerate}
    \item \textbf{Algorithmic:} We introduce Search for Natural Adversarial Features Using Embeddings (SNAFUE) as a tool for scalable human oversight. 
    \item \textbf{Diagnostic:} We apply SNAFUE by red-teaming an image classifier. We demonstrate that it automatedly identifies weaknesses due to natural features that are uniquely human-interpretable. 
\end{enumerate}

Meanwhile, in concurrent work \cite{casper2023benchmarking}, we compare SNAFUE to other interpretability tools by using them to help humans rediscover trojans in a network. We find that SNAFUE offers an effective and unique tool for helping humans interpret and debug DNNs.
Code is available at \href{https://github.com/thestephencasper/snafue}{https://github.com/thestephencasper/snafue}.

\section{Related Work} \label{sec:related_work}

\textbf{Describable Synthetic Adversarial Attacks:} Conventional adversarial attacks are effective but difficult for a human to interpret.
They tend to be imperceptible and, when exaggerated, typically appear as random or mildly-textured noise \cite{szegedy2013intriguing, goodfellow2014explaining}.
Thus, from a human-interpretability perspective, they demonstrate little aside from how the network can be vulnerable to this specific class of perturbations. 
Some works have used perturbations inside the latents of image generators to synthesize more describable synthetic attacks \cite{liu2018beyond, samangouei2018explaingan, song2018constructing, joshi2018xgems, joshi2019semantic, singla2019explanation, hu2021naturalistic, wang2020generating}.
These works however, have not focused on interpretability and only studied small networks trained on simple datasets (MNIST \cite{lecun2010mnist}, Fashion MNIST \cite{xiao2017fashion}, SVHN \cite{netzer2011reading}, CelebA \cite{liu2015faceattributes}, BDD \cite{yu2018bdd100k}, INRIA \cite{dalal2005histograms}, and MPII \cite{andriluka20142d}).
In the trojan detection literature, some methods have aimed to reconstruct trojan features using regularization and transformations.
Thus far, however, they have been limited to recovering small, few-pixel triggers while failing to reconstruct triggers that take the form of larger objects \cite{wang2019neural, guo2019tabor}. 

\textbf{Natural Adversarial Features:} Several approaches have been used for discovering natural adversarial features.
One is to analyze examples in a test set that a DNN mishandles \cite{hendrycks2021natural, eyuboglu2022domino, jain2022distilling}, but this limits the search for weaknesses to a fixed dataset and cannot be used for discovering adversarial \emph{combinations} of features.
Another approach is to search for failures over an easily-describable set of perturbations \cite{geirhos2018imagenet, leclerc20213db, stimberg2023benchmarking}, but this requires performing a zero-order search over a fixed set of changes.

\textbf{Copy Paste Attacks:} Copy/paste attacks have been a growing topic of interest and offer another method for studying natural adversarial features. 
Some interpretability tools have been used to design copy/paste adversarial examples including feature-visualization \cite{carter2019activation} and methods based on network dissection \cite{bau2017dissection, mu2020compositional, hernandez2022natural}.
Our approach is related to that of \cite{casper2022robust} who introduce robust feature level adversarial patches and use them for interpreting DNNs and designing copy-paste attacks.
However, copy/paste attacks from \cite{carter2019activation, mu2020compositional, hernandez2022natural, casper2022robust} have been limited to simple proofs of concept with manually-designed copy/paste attacks. 
They also required a human process of interpretation, trial, and error in the loop. 
We build off of these with SNAFUE which is the first method that identify adversarial combinations of natural features for vision models in a way that is (1) not restricted to a fixed set of transformations or a limited set of source and target classes and (2) efficiently automatable.

\section{Methods} \label{sec:methods}

\figref{fig:fig1} outlines our approach for finding copy/paste adversaries for image classification. 
For all experiments, we report the \emph{success rate} defined as the proportion of the time that a patched image was classified as the target class minus the proportion of the time the unpatched natural image was. 
Additional details are in \apref{app:details}.

\medskip

\noindent \textbf{Synthetic adversarial patches:} First, we create synthetic robust feature level adversarial patches as in \cite{casper2022robust} by perturbing the latent activations of a BigGAN \cite{brock2018large} generator. 
The synthetic adversarial patches were trained to cause any source class image to be misclassified as the target regardless of the insertion location in the source image. 
% We synthesized 30 and selected the $M=10$ that increased the target network's post-softmax confidence in the target class the most. 

\medskip

\noindent \textbf{Candidate patches:} Patches for SNAFUE can come from any source and do not need labels. Features do not necessarily have to be natural and could, for example, be procedurally generated. Here, we used a total of $N=$ 265,457 natural images from five sources: the ImageNet validation set \cite{russakovsky2015imagenet} (50,000) TinyImageNet \cite{le2015tiny} (100,000), OpenSurfaces \cite{bell13opensurfaces} (57,500), the non OpenSurfaces images from Broden \cite{bau2017dissection} (37,953), plus four trojan triggers (4) (see \secref{sec:experiments}). 

\begin{figure}[t!]
    \centering
    \includegraphics[width=0.9\linewidth]{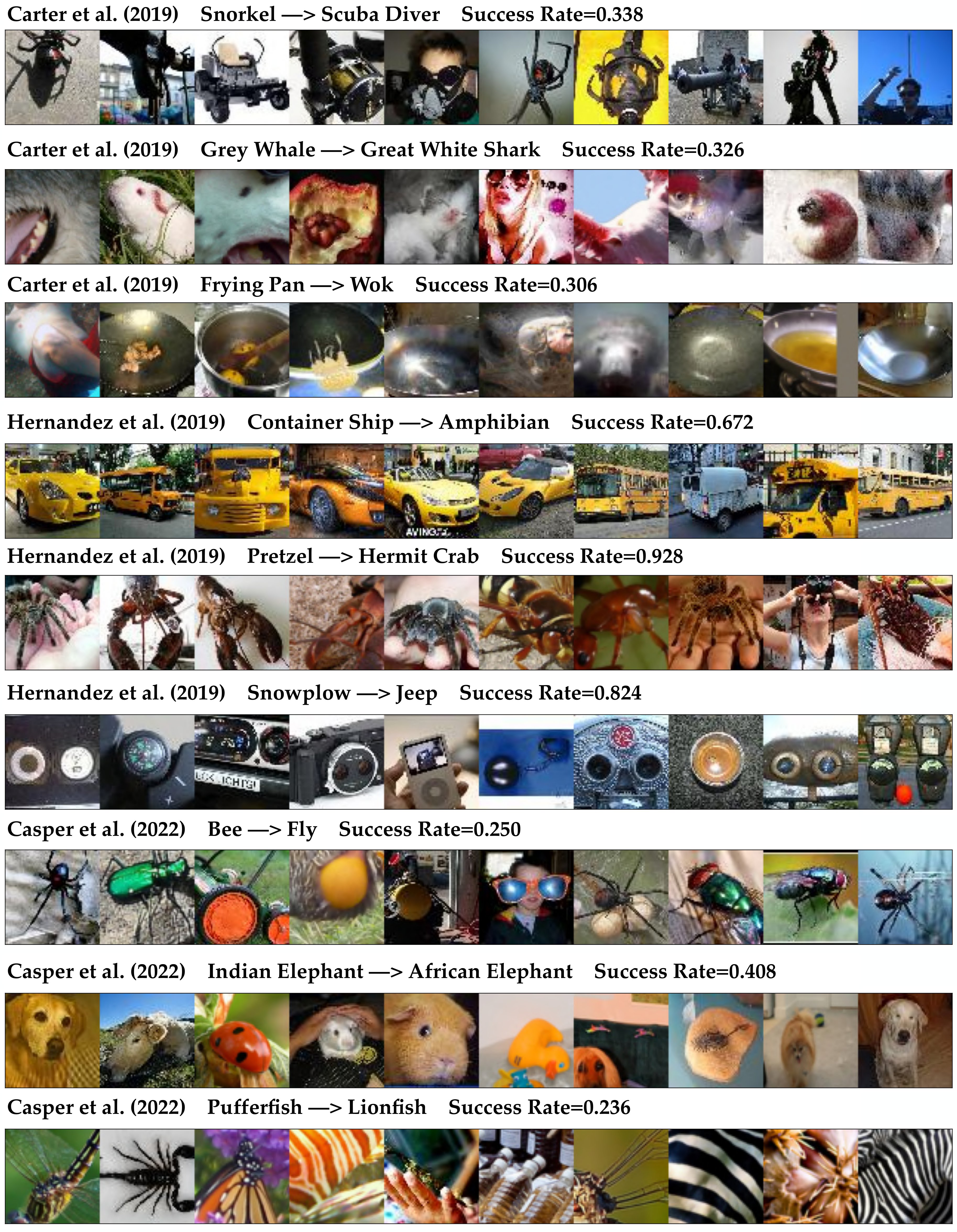}
    \caption{Our automated replications of all 9 prior examples of ImageNet copy/paste attacks of which we are aware from \cite{carter2019activation, hernandez2022natural} and \cite{casper2022robust}. Each set of images is labeled \texttt{source class} $\to$ \texttt{target class}. Each row of 10 patches is labeled with their mean success rate. }
    \label{fig:replication}
\end{figure}

\medskip

\noindent \textbf{Embeddings:} We used the $N=$ 265,457 natural patches along with $M=10$ adversarial patches, and passed them through the target network to get an $L$-dimensional embedding of each using the post-ReLU latents from the penultimate (avgpooling) layer of the target network.
The result was a nonnegative $N \times L$ matrix $U$ of natural patch embeddings and a $M \times L$ matrix $V$ of adversarial patch embeddings. 
A different $V$ must be computed for each attack, but $U$ only needs to be computed once.
This plus the fact that embedding the natural patches does not require insertion into a set of source images makes SNAFUE much more efficient than a brute-force search.
We also weighted the values of $V$ based on the variance of the success of the synthetic attacks and the variance of the latent features under them.
Details are in \apref{app:details}.

\begin{figure*}[t!]
    \centering
    \includegraphics[width=0.95\linewidth]{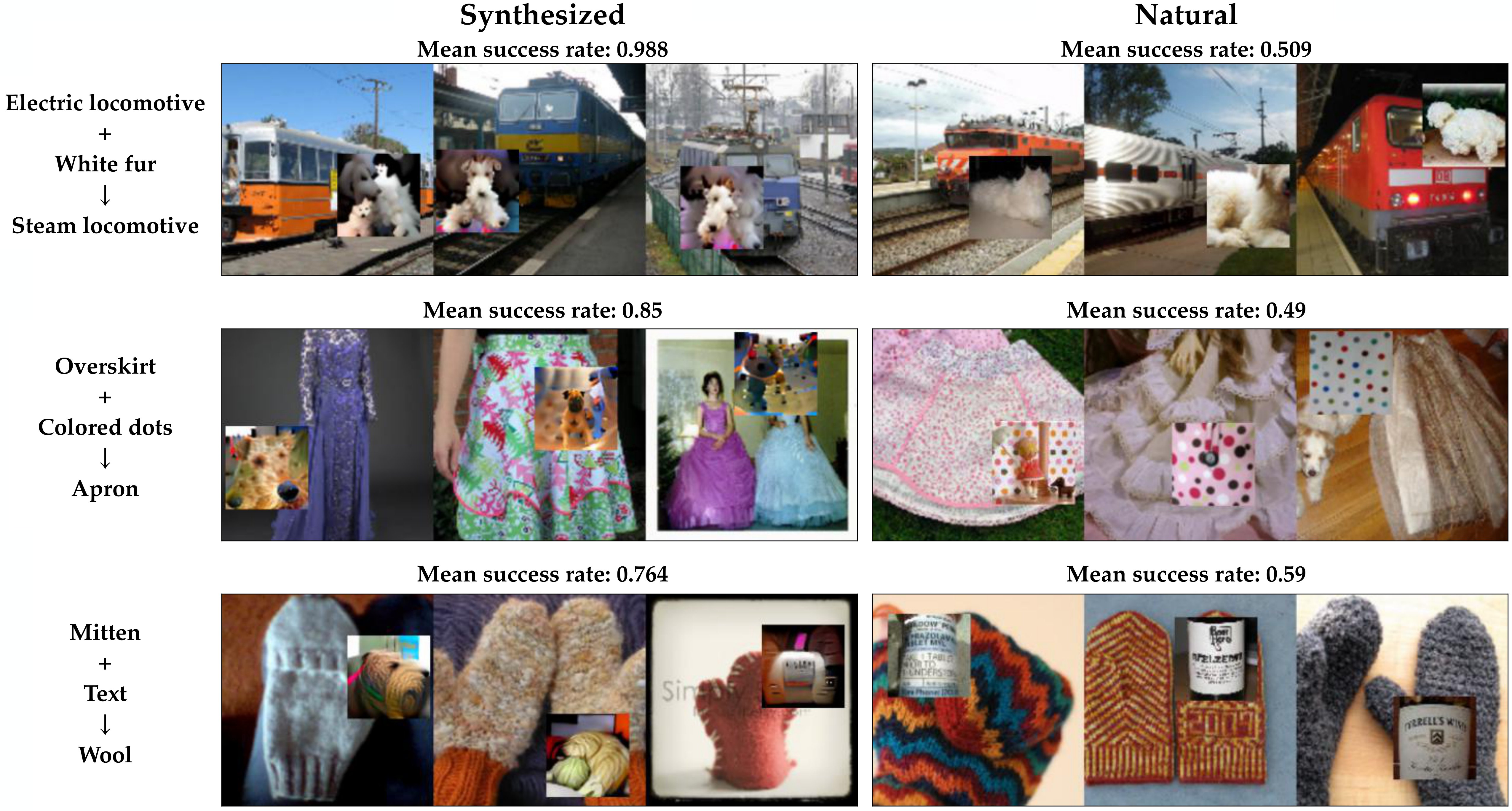}\\
    \bigskip
    \includegraphics[width=0.6\linewidth]{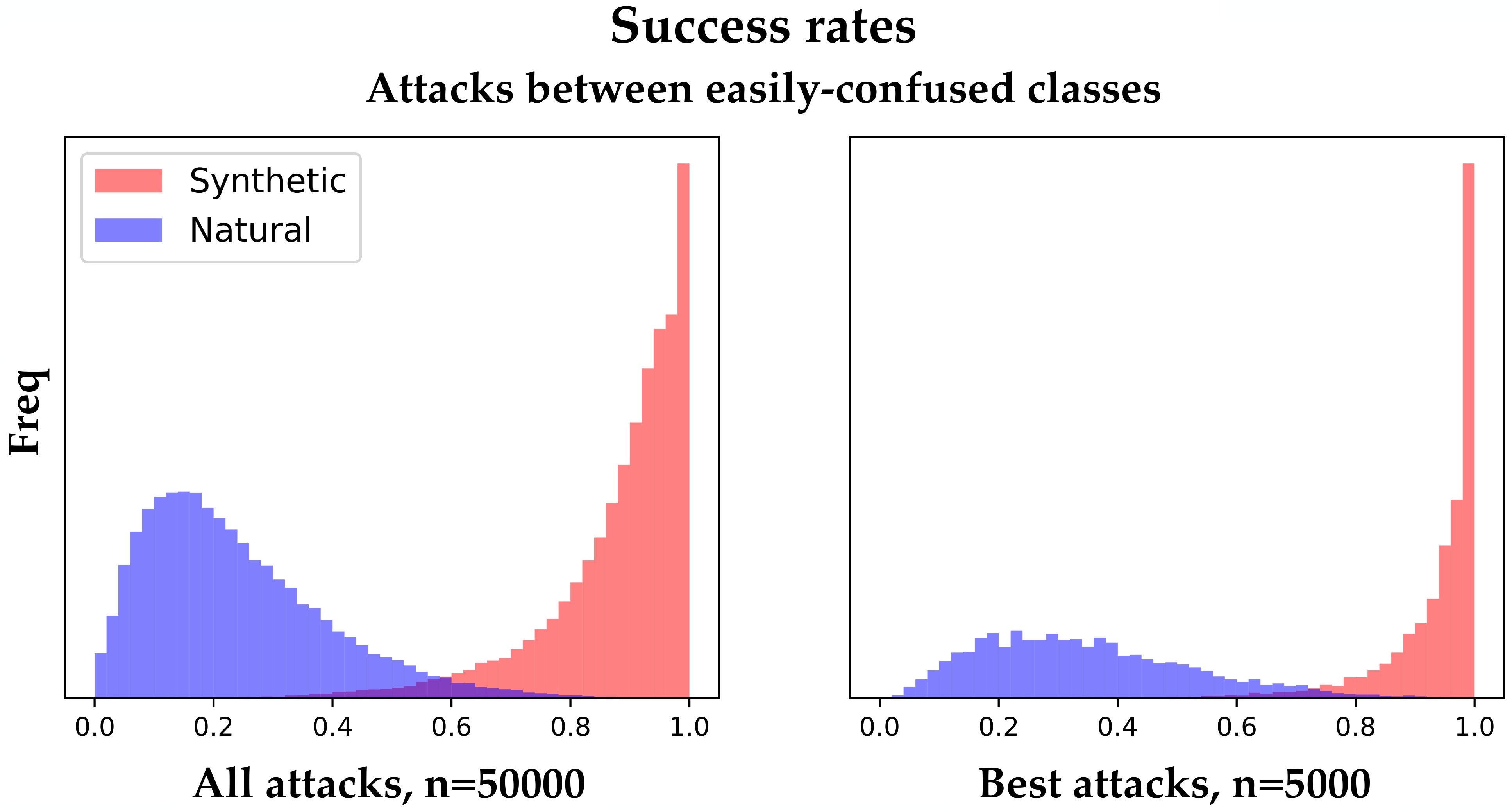}
    \caption{(Top) Examples of copy/paste attacks between similar source/target classes. Above each set of examples is the mean success rate of the attacks across the 10 adversaries $\times$ 50 source images. (Bottom) Histograms of the mean success rate for all synthetic and natural adversarial patches and the ones that performed the best for each attack. Labels for the adversarial features (e.g. ``white fur'') are human-produced.}
    \label{fig:breadth}
\end{figure*}

\medskip

\noindent \textbf{Selecting natural patches:} We then obtained the $N \times M$ matrix $S$ of cosine similarities between $U$ and $V$.
We took the $K'=300$ patches that had the highest similarity to \emph{any} of the synthetic images, excluding ones whose classifications from the target network included the target class in the top 10 classes.
Finally, we evaluated all $K'$ natural patches under random insertion locations over all 50 source images from the validation set and subsampled the $K=10$ natural patches that increased the target network's post-softmax confidence in the target class the most.
Screening the $K'$ natural patches for the best 10 caused only a marginal increase in computational overhead. 
The method was mainly bottlenecked by the cost of training the synthetic adversarial patches (for 64 batches of 32 insertions each). 

\medskip

\noindent \textbf{Automation of interpretations:} Unlike previous proofs of concept, SNAFUE does not rely on a human in the loop.
However, we still rely on a human after the loop for analyzing the final results and making a final interpretation by default.
To test how easily this may be automatable, we test a method for turning many images into a single caption in \apref{app:humans}.

% \medskip

% \noindent \textbf{Screening:} In \apref{app:syn_nat}, we test how the performance of the final copy/paste attacks varies as a function of the total number $M'$ of synthetic adversaries created and the number $K'$ of natural adversaries screened while selecting the best $M$ synthetic ones and $K$ natural ones to sue. 
% We vary the total number $M'$ of synthetic adversaries created from 4 to 64 and found no evidence of performance improving by increasing the number of synthetic patches and selecting the best $M=\min(M', 10)$ ones.
% We then varied the number of natural patches $K'$ from 10 to 2560 while holding the number of selected ones $K=10$ to be constant and found clear but diminishing improvements as $K'$ increased.

\section{Experiments} \label{sec:experiments}

\noindent \textbf{Replicating previous ImageNet copy/paste attacks without human involvement.} First, we set out to replicate \emph{all} known successful ImageNet copy/paste attacks from previous works without any human involvement. 
To our knowledge, there are 9 such attacks, 3 each from \cite{carter2019activation}, \cite{hernandez2022natural}\footnote{The attacks presented in \cite{hernandez2022natural} were not universal within a source class and were only developed for a single source image each. When replicating their results, we use the same single sources. When replicating attacks from the other two works, we train and test the attacks as source class-universal ones.} and \cite{casper2022robust}.\footnote{\cite{casper2022robust} test a fourth attack involving patches making traffic lights appear as flies, the examples they identified were not successful at causing targeted misclassification.}\footnote{\cite{mu2020compositional} also test copy paste attacks, but not on ImageNet networks}
We used SNAFUE to find 10 natural patches for all 9 attacks. 
\figref{fig:replication} shows the results.
In all cases, we are able to find successful natural adversarial patches.
We also find in most cases that we find similar adversarial features to the ones identified in the prior works. 
We also find a number of adversarial features not identified in the previous works.

\begin{figure*}
    \centering
    \includegraphics[width=0.95\linewidth]{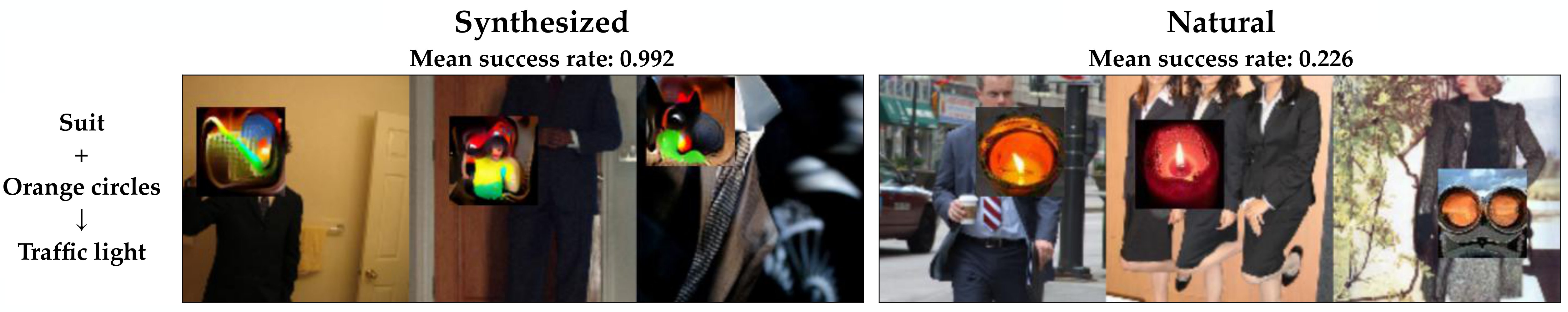}\\
    \bigskip
    \includegraphics[width=0.6\linewidth]{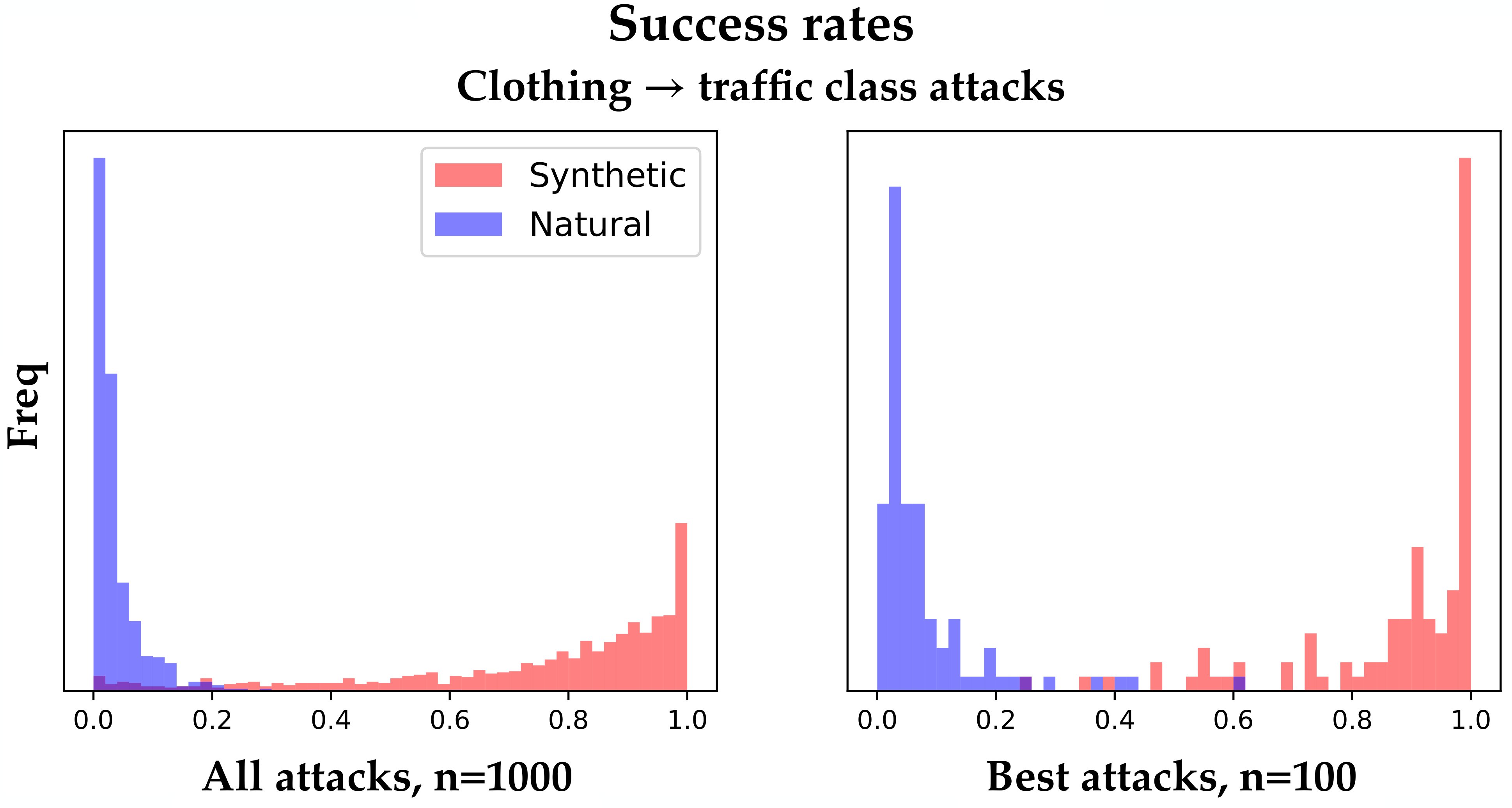}
    \caption{(Top) Examples from our most successful copy/paste attack using a clothing source and a traffic target. The mean success rate of the attacks across 10 adversaries $\times$ 50 source images is shown above each example. (Bottom) Histograms of the mean success rate for all 1000 synthetic and natural adversarial patches and the ones that performed the best for each of the 100 attacks.}
    \label{fig:clothing_to_traffic}
\end{figure*}

\noindent \textbf{SNAFUE is scalable and effective between similar classes.} There are many natural visual features that image classifiers may encounter and many more possible combinations thereof, so it is important that tools for interpretability and diagnostics with natural features are scalable. 
Here, we test the scalability and effectiveness of SNAFUE with a broad search for vulnerabilities. 
Based on prior proofs of concept \cite{carter2019activation, mu2020compositional, hernandez2022natural, casper2022robust} copy/paste attacks tend to be much easier to create when the source and target class are related (see \figref{fig:replication}). 
To choose similar source/target pairs, we computed the confusion matrix $C$ for the target network with $0 \le C_{ij} \le 1$ giving the mean post-softmax confidence on class $j$ that the network assigned to validation images of label $i$.
Then for each of the 1,000 ImageNet classes, we conducted 5 attacks using that class as the source and each of its most confused 5 classes as targets. 
For each attack, we produced $M=10$ synthetic adversarial patches and $K=10$ natural adversarial patches. 
\figref{fig:nat_examples1} and \figref{fig:breadth} show examples from these attacks with many additional examples in Appendix \figref{fig:nat_examples2}.
Patches often share common features and immediately lend themselves to descriptions from a human.
In \apref{app:humans}, we find that these patches also often lend themselves to common descriptions from neural captioning models.

At the bottom of \figref{fig:breadth}, are histograms for the mean attack success rate for all patches and for the best patches (out of 10) for each attack. 
The synthetic feature-level adversaries were generally highly successful, and the the natural patches were also successful a significant proportion of the time.
In this experiment, 3,451 (6.9\%) out of the 50,000 total natural images from all attacks were at least 50\% successful at being \emph{targeted} adversarial patches under random insertion locations into random image of the source class. 
This compares to a 10.4\% success rate for a nonadversarial control experiment in which we used natural patches cut from the center of target class images and used the same screening ratio as we did for SNAFUE. 
Meanwhile, 963 (19.5\%) of the 5,000 best natural images were at least 50\% successful, and interestingly, in \emph{all but one} of the 5,000 total source/target class pairs, at least one natural image was found which fooled the classifier as a targeted attack for at least one source image.

\medskip

\noindent \textbf{Copy/paste attacks between dissimilar classes are possible but more challenging.}
In some cases, the ability to robustly distinguish between similar classes may be crucial.
For example, it is important for autonomous vehicles to effectively tell red and yellow traffic lights apart.
But studying how easily networks can be made to mistake an image for \emph{arbitrary} target classes is of broader general interest. 
While synthetic adversarial attacks often work between arbitrary source/target classes, to the best of our knowledge, there are no successful examples from any previous works of class-universal copy/paste attacks.

We chose to examine the practical problem of understanding how vision systems in vehicles may fail to detect pedestrians \cite{ntsb2018collision} because it provides an example where failures due to novel combinations of natural features could realistically pose safety hazards. 
To test attacks between dissimilar classes, we chose 10 ImageNet classes of clothing items (which frequently co-occur with humans) and 10 of traffic-related objects.\footnote{\{academic gown, apron, bikini, cardigan, jean, jersey, maillot, suit, sweatshirt, trenchcoat\} $\times$ \{fire engine, garbage truck, racer, sports car, streetcar, tow truck, trailer truck, trolleybus, street sign, traffic light\}} 
We conducted 100 total attacks with SNAFUE using each clothing source and traffic target. \figref{fig:clothing_to_traffic} shows these results.
Outcomes were mixed.

On one hand, while the synthetic adversarial patches were usually successful on more than 50\% of source images, the natural ones were usually not.
Only one out of the 1,000 total natural patches (the leftmost natural patch in \figref{fig:clothing_to_traffic}) succeeded for at least 50\% of source class images.
This suggests a limitation of either SNAFUE or of copy/paste attacks in general for targeted attacks between unrelated source and target classes. 
On the other hand, 54\% of the natural adversarial patches were successful for at least one source image, and such a natural patch was identified for 87 of all 100 source/target class pairs. 

\medskip

\noindent \textbf{SNAFUE is unique and relatively effective compared to other interpretability/diagnostic tools:} Finally, we test how SNAFUE compares to related tools meant to help humans better understand and diagnose bugs in models. 
In concurrent work \cite{casper2023benchmarking}, we evaluate interpretability tools for DNNs based on how effective they are at helping humans identify trojans that have been implanted into DNNs. 
We test SNAFUE against 8 other interpretability tools for DNNs based on feature synthesis/search. 
We find that SNAFUE and robust feature level adversaries \cite{casper2022robust} are the most successful overall and that combinations of methods are more helpful than individual ones.

\section{Discussion and Broader Impact} \label{sec:discussion}

\textbf{Implications for scalable human oversight.}
Having effective diagnostic tools to identify problems with models is important for trustworthy AI.
The most common way to evaluate a model is with a test set.
But good testing performance does not imply that a system will generalize safely in deployment.
Test sets do not typically reveal failures such as spurious features, out of distribution inputs, and adversarial vulnerabilities.
Thus, it is important to have scalable tools that allow \emph{humans} to exercise effective oversight over deep neural networks. 

Interpretability tools are useful for building more trustworthy AI because of the role they can play in helping humans exercise oversight. 
But many techniques from the literature suffer from limitations including a lack of scalability and usefulness for identifying \emph{novel} flaws.
This has led to criticism, with a number of works noting that few interpretability tools are used by practitioners in real applications \cite{doshivelez2017towards, lipton2018mythos, miller2019explanation, krishnan2020against, raukur2022toward}.
Toward practical methods to find weaknesses in DNNs, we introduce SNAFUE as an automated method for finding natural adversarial features.

\textbf{SNAFUE identifies distinct types of problems.} 
In some cases, networks may learn flawed solutions because they are given the wrong learning objective while in other cases, they may fail to converge to a desirable solution even with the correct objective \cite{hubinger2019risks}.
SNAFUE can discover both types of issues. 
In some cases, it discovers failures that result from dataset biases. 
Examples include when it identifies that cats make envelopes misclassified as cartons or that young children make bicycles-built-for-two misclassified as tricycles (\figref{fig:nat_examples1} rows 1-2). 
In other cases, SNAFUE identifies failures that result from the particular representations a model learns, presumably due to equivalence classes in the DNN's representations. 
Examples include equating black and white birds with killer whales, parallel lines with spatulas, and red/orange cars with fiddler crabs (\figref{fig:nat_examples1} rows 3-5).

\textbf{Limitations.} We find that it scales well and can easily identify hundreds of sets of copy/paste vulnerabilities that are very easy for a human to interpret and describe. 
However, we also find limitations including how SNAFUE is less effective for dissimilar source and target classes.
In \apref{app:failures}, we catrogize and discuss different types of failures with SNAFUE.

\textbf{Three directions for future work.}
\begin{enumerate}
    \item \textbf{Diagnostics in the wild:} Vision datasets are full of biases, including harmful ones involving human demographic groups \cite{fabbrizzi2022survey}. A compelling use of SNAFUE and similar techniques could be for discovering these in deployed systems. This could be valuable for exploring the practical relevance of diagnostic tools. 
    \item \textbf{Debugging:} In addition to is use for interpretability, SNAFUE also produces adversarial data which can be used for adversarial training or probing the network to guide targeted procedural edits. Correcting vulnerabilities to copy/paste attacks could be useful applications or tests for model editing tools (e.g. \cite{ghorbani2020shapley, dai2021knowledge, wong2021leveraging, meng2022locating}). In the real world, vision systems often fail due to distractor features, atypical contexts, and occlusion \cite{hendrycks2021natural}. Debugging with copy/paste attacks may be well-equipped to address these failures. 
    \item \textbf{NLP:} Using a version of SNAFUE in natural language processing could be helpful for identifying natural phrases that could cause language models to fail. This could be valuable for language models because the discrete nature of their inputs makes it difficult to use gradient based methods to construct adversarial perturbations in \emph{input space}. However, SNAFUE could be used to produce interpretable adversaries from synthetic adversarial insertions to the \emph{latents}. Other recent works have aimed to generate adversarial triggers for language models that appear as natural language phrases. However, these often depend on either using reinforcement learning to train language generators which can be unstable and computationally expensive \cite{perez2022red} or on using humans in the loop \cite{ziegler2022adversarial}. Because SNAFUE is automated and can work flexibly with any dataset of candidate features, it may offer a competitive alternative to existing tools. 
\end{enumerate}

All of the proposals for building safe AI outlined in \cite{hubinger2020overview} explicitly call for adversarial robustness and/or oversight via interpretability tools.
Finding and fixing bugs in advanced AI systems will hinge on interpretability, adversaries, and adversarial training.
Consequently, continued work toward scalable techniques for interpretability and diagnostics will be important for safer AI.

\section*{Acknowledgments}
We thank Rui-Jie Yew and Phillip Christoffersen for feedback and Evan Hernandez for discussing how to best replicate results from \cite{hernandez2022natural}. Stephen Casper's work was supported by the Future of Life Institute and Kaivalya Hariharan's work was supported by the Open Philanthropy Project.

\bibliographystyle{plain}
\bibliography{bibliography}

\begin{thebibliography}{10}

\bibitem{andriluka20142d}
Mykhaylo Andriluka, Leonid Pishchulin, Peter Gehler, and Bernt Schiele.
\newblock 2d human pose estimation: New benchmark and state of the art
  analysis.
\newblock In {\em Proceedings of the IEEE Conference on computer Vision and
  Pattern Recognition}, pages 3686--3693, 2014.

\bibitem{apruzzese2022real}
Giovanni Apruzzese, Hyrum~S Anderson, Savino Dambra, David Freeman, Fabio
  Pierazzi, and Kevin~A Roundy.
\newblock " real attackers don't compute gradients": Bridging the gap between
  adversarial ml research and practice.
\newblock {\em arXiv preprint arXiv:2212.14315}, 2022.

\bibitem{bau2017dissection}
David Bau, Bolei Zhou, Aditya Khosla, Aude Oliva, and Antonio Torralba.
\newblock Network dissection: Quantifying interpretability of deep visual
  representations, 2017.

\bibitem{bell13opensurfaces}
Sean Bell, Paul Upchurch, Noah Snavely, and Kavita Bala.
\newblock Open{S}urfaces: A richly annotated catalog of surface appearance.
\newblock {\em ACM Trans. on Graphics (SIGGRAPH)}, 32(4), 2013.

\bibitem{borowski2020exemplary}
Judy Borowski, Roland~S Zimmermann, Judith Schepers, Robert Geirhos, Thomas~SA
  Wallis, Matthias Bethge, and Wieland Brendel.
\newblock Exemplary natural images explain cnn activations better than
  state-of-the-art feature visualization.
\newblock {\em arXiv preprint arXiv:2010.12606}, 2020.

\bibitem{brock2018large}
Andrew Brock, Jeff Donahue, and Karen Simonyan.
\newblock Large scale gan training for high fidelity natural image synthesis.
\newblock {\em arXiv preprint arXiv:1809.11096}, 2018.

\bibitem{brown2017adversarial}
Tom~B Brown, Dandelion Man{\'e}, Aurko Roy, Mart{\'\i}n Abadi, and Justin
  Gilmer.
\newblock Adversarial patch.
\newblock {\em arXiv preprint arXiv:1712.09665}, 2017.

\bibitem{carter2019activation}
Shan Carter, Zan Armstrong, Ludwig Schubert, Ian Johnson, and Chris Olah.
\newblock Activation atlas.
\newblock {\em Distill}, 4(3):e15, 2019.

\bibitem{casper2023benchmarking}
Stephen Casper, Yuxiao Li, Jiawei Li, Tong Bu, Kevin Zhang, and Dylan
  Hadfield-Menell.
\newblock Benchmarking interpretability tools for deep neural networks.
\newblock {\em arXiv preprint arXiv:2302.10894}, 2023.

\bibitem{casper2022robust}
Stephen Casper, Max Nadeau, and Gabriel Kreiman.
\newblock Robust feature-level adversaries are interpretability tools.
\newblock {\em CoRR}, abs/2110.03605, 2021.

\bibitem{dai2021knowledge}
Damai Dai, Li~Dong, Yaru Hao, Zhifang Sui, and Furu Wei.
\newblock Knowledge neurons in pretrained transformers.
\newblock {\em arXiv preprint arXiv:2104.08696}, 2021.

\bibitem{dalal2005histograms}
Navneet Dalal and Bill Triggs.
\newblock Histograms of oriented gradients for human detection.
\newblock In {\em 2005 IEEE computer society conference on computer vision and
  pattern recognition (CVPR'05)}, volume~1, pages 886--893. Ieee, 2005.

\bibitem{dong2017towards}
Yinpeng Dong, Hang Su, Jun Zhu, and Fan Bao.
\newblock Towards interpretable deep neural networks by leveraging adversarial
  examples.
\newblock {\em arXiv preprint arXiv:1708.05493}, 2017.

\bibitem{doshivelez2017towards}
Finale Doshi-Velez and Been Kim.
\newblock Towards a rigorous science of interpretable machine learning, 2017.

\bibitem{eyuboglu2022domino}
Sabri Eyuboglu, Maya Varma, Khaled Saab, Jean-Benoit Delbrouck, Christopher
  Lee-Messer, Jared Dunnmon, James Zou, and Christopher R{\'e}.
\newblock Domino: Discovering systematic errors with cross-modal embeddings.
\newblock {\em arXiv preprint arXiv:2203.14960}, 2022.

\bibitem{fabbrizzi2022survey}
Simone Fabbrizzi, Symeon Papadopoulos, Eirini Ntoutsi, and Ioannis
  Kompatsiaris.
\newblock A survey on bias in visual datasets.
\newblock {\em Computer Vision and Image Understanding}, 223:103552, 2022.

\bibitem{geirhos2018imagenet}
Robert Geirhos, Patricia Rubisch, Claudio Michaelis, Matthias Bethge, Felix~A
  Wichmann, and Wieland Brendel.
\newblock Imagenet-trained cnns are biased towards texture; increasing shape
  bias improves accuracy and robustness.
\newblock {\em arXiv preprint arXiv:1811.12231}, 2018.

\bibitem{ghorbani2020shapley}
Amirata Ghorbani and James Zou.
\newblock Neuron shapley: Discovering the responsible neurons, 2020.

\bibitem{goodfellow2014explaining}
Ian~J Goodfellow, Jonathon Shlens, and Christian Szegedy.
\newblock Explaining and harnessing adversarial examples.
\newblock {\em arXiv preprint arXiv:1412.6572}, 2014.

\bibitem{guo2019tabor}
Wenbo Guo, Lun Wang, Xinyu Xing, Min Du, and Dawn Song.
\newblock Tabor: A highly accurate approach to inspecting and restoring trojan
  backdoors in ai systems.
\newblock {\em arXiv preprint arXiv:1908.01763}, 2019.

\bibitem{he2016deep}
Kaiming He, Xiangyu Zhang, Shaoqing Ren, and Jian Sun.
\newblock Deep residual learning for image recognition.
\newblock In {\em Proceedings of the IEEE conference on computer vision and
  pattern recognition}, pages 770--778, 2016.

\bibitem{hendrycks2021natural}
Dan Hendrycks, Kevin Zhao, Steven Basart, Jacob Steinhardt, and Dawn Song.
\newblock Natural adversarial examples.
\newblock In {\em Proceedings of the IEEE/CVF Conference on Computer Vision and
  Pattern Recognition}, pages 15262--15271, 2021.

\bibitem{hernandez2022natural}
Evan Hernandez, Sarah Schwettmann, David Bau, Teona Bagashvili, Antonio
  Torralba, and Jacob Andreas.
\newblock Natural language descriptions of deep visual features.
\newblock {\em arXiv preprint arXiv:2201.11114}, 2022.

\bibitem{hu2021naturalistic}
Yu-Chih-Tuan Hu, Bo-Han Kung, Daniel~Stanley Tan, Jun-Cheng Chen, Kai-Lung Hua,
  and Wen-Huang Cheng.
\newblock Naturalistic physical adversarial patch for object detectors.
\newblock In {\em Proceedings of the IEEE/CVF International Conference on
  Computer Vision}, pages 7848--7857, 2021.

\bibitem{hubinger2020overview}
Evan Hubinger.
\newblock An overview of 11 proposals for building safe advanced ai.
\newblock {\em arXiv preprint arXiv:2012.07532}, 2020.

\bibitem{hubinger2019risks}
Evan Hubinger, Chris van Merwijk, Vladimir Mikulik, Joar Skalse, and Scott
  Garrabrant.
\newblock Risks from learned optimization in advanced machine learning systems.
\newblock {\em arXiv preprint arXiv:1906.01820}, 2019.

\bibitem{ilyas2019adversarial}
Andrew Ilyas, Shibani Santurkar, Dimitris Tsipras, Logan Engstrom, Brandon
  Tran, and Aleksander Madry.
\newblock Adversarial examples are not bugs, they are features.
\newblock {\em Advances in neural information processing systems}, 32, 2019.

\bibitem{jain2022distilling}
Saachi Jain, Hannah Lawrence, Ankur Moitra, and Aleksander Madry.
\newblock Distilling model failures as directions in latent space, 2022.

\bibitem{joshi2019semantic}
Ameya Joshi, Amitangshu Mukherjee, Soumik Sarkar, and Chinmay Hegde.
\newblock Semantic adversarial attacks: Parametric transformations that fool
  deep classifiers.
\newblock In {\em Proceedings of the IEEE/CVF International Conference on
  Computer Vision}, pages 4773--4783, 2019.

\bibitem{joshi2018xgems}
Shalmali Joshi, Oluwasanmi Koyejo, Been Kim, and Joydeep Ghosh.
\newblock xgems: Generating examplars to explain black-box models.
\newblock {\em arXiv preprint arXiv:1806.08867}, 2018.

\bibitem{krishnan2020against}
Maya Krishnan.
\newblock Against interpretability: a critical examination of the
  interpretability problem in machine learning.
\newblock {\em Philosophy \& Technology}, 33(3):487--502, 2020.

\bibitem{le2015tiny}
Ya~Le and Xuan Yang.
\newblock Tiny imagenet visual recognition challenge.
\newblock {\em CS 231N}, 7(7):3, 2015.

\bibitem{leclerc20213db}
Guillaume Leclerc, Hadi Salman, Andrew Ilyas, Sai Vemprala, Logan Engstrom,
  Vibhav Vineet, Kai Xiao, Pengchuan Zhang, Shibani Santurkar, Greg Yang,
  et~al.
\newblock 3db: A framework for debugging computer vision models.
\newblock {\em arXiv preprint arXiv:2106.03805}, 2021.

\bibitem{lecun2010mnist}
Yann LeCun, Corinna Cortes, and CJ~Burges.
\newblock Mnist handwritten digit database.
\newblock {\em ATT Labs [Online]. Available: http://yann.lecun.com/exdb/mnist},
  2, 2010.

\bibitem{li2022blip}
Junnan Li, Dongxu Li, Caiming Xiong, and Steven Hoi.
\newblock Blip: Bootstrapping language-image pre-training for unified
  vision-language understanding and generation.
\newblock In {\em ICML}, 2022.

\bibitem{lipton2018mythos}
Zachary~C Lipton.
\newblock The mythos of model interpretability: In machine learning, the
  concept of interpretability is both important and slippery.
\newblock {\em Queue}, 16(3):31--57, 2018.

\bibitem{liu2018beyond}
Hsueh-Ti~Derek Liu, Michael Tao, Chun-Liang Li, Derek Nowrouzezahrai, and Alec
  Jacobson.
\newblock Beyond pixel norm-balls: Parametric adversaries using an analytically
  differentiable renderer.
\newblock {\em arXiv preprint arXiv:1808.02651}, 2018.

\bibitem{liu2015faceattributes}
Ziwei Liu, Ping Luo, Xiaogang Wang, and Xiaoou Tang.
\newblock Deep learning face attributes in the wild.
\newblock In {\em Proceedings of International Conference on Computer Vision
  (ICCV)}, December 2015.

\bibitem{meng2022locating}
Kevin Meng, David Bau, Alex Andonian, and Yonatan Belinkov.
\newblock Locating and editing factual associations in gpt.
\newblock {\em arXiv preprint arXiv:2202.05262}, 2022.

\bibitem{miller2019explanation}
Tim Miller.
\newblock Explanation in artificial intelligence: Insights from the social
  sciences.
\newblock {\em Artificial intelligence}, 267:1--38, 2019.

\bibitem{mu2020compositional}
Jesse Mu and Jacob Andreas.
\newblock Compositional explanations of neurons.
\newblock {\em Advances in Neural Information Processing Systems},
  33:17153--17163, 2020.

\bibitem{netzer2011reading}
Yuval Netzer, Tao Wang, Adam Coates, Alessandro Bissacco, Bo~Wu, and Andrew~Y
  Ng.
\newblock Reading digits in natural images with unsupervised feature learning,
  2011.

\bibitem{ntsb2018collision}
National Transportation Safety~Board NTSB.
\newblock Collision between vehicle controlled by developmental automated
  driving system and pedestrian, 2018.

\bibitem{perez2022red}
Ethan Perez, Saffron Huang, Francis Song, Trevor Cai, Roman Ring, John
  Aslanides, Amelia Glaese, Nat McAleese, and Geoffrey Irving.
\newblock Red teaming language models with language models.
\newblock {\em arXiv preprint arXiv:2202.03286}, 2022.

\bibitem{raukur2022toward}
Tilman R{\"a}ukur, Anson Ho, Stephen Casper, and Dylan Hadfield-Menell.
\newblock Toward transparent ai: A survey on interpreting the inner structures
  of deep neural networks.
\newblock {\em arXiv preprint arXiv:2207.13243}, 2022.

\bibitem{russakovsky2015imagenet}
Olga Russakovsky, Jia Deng, Hao Su, Jonathan Krause, Sanjeev Satheesh, Sean Ma,
  Zhiheng Huang, Andrej Karpathy, Aditya Khosla, Michael Bernstein, et~al.
\newblock Imagenet large scale visual recognition challenge.
\newblock {\em International journal of computer vision}, 115(3):211--252,
  2015.

\bibitem{samangouei2018explaingan}
Pouya Samangouei, Ardavan Saeedi, Liam Nakagawa, and Nathan Silberman.
\newblock Explaingan: Model explanation via decision boundary crossing
  transformations.
\newblock In {\em Proceedings of the European Conference on Computer Vision
  (ECCV)}, pages 666--681, 2018.

\bibitem{schulman2022chatgpt}
J~Schulman, B~Zoph, C~Kim, J~Hilton, J~Menick, J~Weng, JFC Uribe, L~Fedus,
  L~Metz, M~Pokorny, et~al.
\newblock Chatgpt: Optimizing language models for dialogue, 2022.

\bibitem{singla2019explanation}
Sumedha Singla, Brian Pollack, Junxiang Chen, and Kayhan Batmanghelich.
\newblock Explanation by progressive exaggeration.
\newblock {\em arXiv preprint arXiv:1911.00483}, 2019.

\bibitem{song2018constructing}
Yang Song, Rui Shu, Nate Kushman, and Stefano Ermon.
\newblock Constructing unrestricted adversarial examples with generative
  models.
\newblock {\em arXiv preprint arXiv:1805.07894}, 2018.

\bibitem{stimberg2023benchmarking}
Florian Stimberg, Ayan Chakrabarti, Chun-Ta Lu, Hussein Hazimeh, Otilia
  Stretcu, Wei Qiao, Yintao Liu, Merve Kaya, Cyrus Rashtchian, Ariel Fuxman,
  Mehmet Tek, and Sven Gowal.
\newblock Benchmarking robustness to adversarial image obfuscations, 2023.

\bibitem{szegedy2013intriguing}
Christian Szegedy, Wojciech Zaremba, Ilya Sutskever, Joan Bruna, Dumitru Erhan,
  Ian Goodfellow, and Rob Fergus.
\newblock Intriguing properties of neural networks.
\newblock {\em arXiv preprint arXiv:1312.6199}, 2013.

\bibitem{tomsett2018failure}
Richard Tomsett, Amy Widdicombe, Tianwei Xing, Supriyo Chakraborty, Simon
  Julier, Prudhvi Gurram, Raghuveer Rao, and Mani Srivastava.
\newblock Why the failure? how adversarial examples can provide insights for
  interpretable machine learning.
\newblock In {\em 2018 21st International Conference on Information Fusion
  (FUSION)}, pages 838--845. IEEE, 2018.

\bibitem{wang2019neural}
Bolun Wang, Yuanshun Yao, Shawn Shan, Huiying Li, Bimal Viswanath, Haitao
  Zheng, and Ben~Y Zhao.
\newblock Neural cleanse: Identifying and mitigating backdoor attacks in neural
  networks.
\newblock In {\em 2019 IEEE Symposium on Security and Privacy (SP)}, pages
  707--723. IEEE, 2019.

\bibitem{wang2020generating}
Shuo Wang, Shangyu Chen, Tianle Chen, Surya Nepal, Carsten Rudolph, and Marthie
  Grobler.
\newblock Generating semantic adversarial examples via feature manipulation.
\newblock {\em arXiv preprint arXiv:2001.02297}, 2020.

\bibitem{wong2021leveraging}
Eric Wong, Shibani Santurkar, and Aleksander Madry.
\newblock Leveraging sparse linear layers for debuggable deep networks.
\newblock In {\em International Conference on Machine Learning}, pages
  11205--11216. PMLR, 2021.

\bibitem{xiao2017fashion}
Han Xiao, Kashif Rasul, and Roland Vollgraf.
\newblock Fashion-mnist: a novel image dataset for benchmarking machine
  learning algorithms.
\newblock {\em arXiv preprint arXiv:1708.07747}, 2017.

\bibitem{yu2018bdd100k}
Fisher Yu, Wenqi Xian, Yingying Chen, Fangchen Liu, Mike Liao, Vashisht
  Madhavan, and Trevor Darrell.
\newblock Bdd100k: A diverse driving video database with scalable annotation
  tooling.
\newblock {\em arXiv preprint arXiv:1805.04687}, 2(5):6, 2018.

\bibitem{ziegler2022adversarial}
Daniel~M Ziegler, Seraphina Nix, Lawrence Chan, Tim Bauman, Peter
  Schmidt-Nielsen, Tao Lin, Adam Scherlis, Noa Nabeshima, Ben Weinstein-Raun,
  Daniel de~Haas, et~al.
\newblock Adversarial training for high-stakes reliability.
\newblock {\em arXiv preprint arXiv:2205.01663}, 2022.

\end{thebibliography}

\newpage
\appendix
\section{Appendix}

\subsection{Additional Methodological Details} \label{app:details}

\noindent \textbf{Network and data:} We attack a ResNet18 \cite{he2016deep} trained on ImageNet \cite{russakovsky2015imagenet}. 

\noindent \textbf{Details on synthetic patches:} As in \cite{casper2022robust}, we optimized these patches under transformation using an auxiliary classifier to regularize them to not appear like the target class.
Unlike \cite{casper2022robust}, we do not use a GAN discriminator for regularization or use an auxiliary classifier to regularize for realistic-looking patches.
Also in contrast with \cite{casper2022robust}, we perturbed the inputs to the generator in addition to its internal activations in a certain layer because we found that it produced improved adversarial patches.

\medskip

\noindent \textbf{Image and patch scaling:} All synthetic patches were parameterized as $64 \times 64$ images. Each was trained under transformations including random resizing. 
Similarly, all natural patches were a uniform resolution of $64 \times 64$ pixels.
All adversarial patches were tested by resizing them to $100 \times 100$ and inserting them into $256 \times 256$ source images at random locations.  

\medskip

\noindent \textbf{Weighting:} To reduce the influence of embedding features that vary widely across the adversarial patches, we apply an $L$-dimensional elementwise mask $w$ to the embedding in each row of $V$ with weights 

$$
w_j = \begin{cases}
0 & \textrm{if } \textrm{cv}_i(V_{ij}) > 1\\
1 - \textrm{cv}_i(V_{ij}) & \textrm{else}
\end{cases}
$$

\noindent where $\textrm{cv}_i(V_{ij})$ is the coefficient of variation over the $j$'th column of $V$, with $\mu_j = \frac{1}{M}\sum_i V_{ij} \ge 0$ and  $\textrm{cv}_i(V_{ij}) = $ $\frac{\sqrt{\frac{1}{M-1}\sum_i (V_{ij} - \mu_j)^2}}{\mu_j + \epsilon}$ for some small positive $\epsilon$.

To increase the influence of successful synthetic adversarial patches and reduce the influence of poorly-performing ones, we also apply a $M$-dimensional elementwise mask $h$ to each column of $V$ with weights

$$h_i = \frac{\delta_i -  \delta_{\min}}{\delta_{\max} - \delta_{\min}}$$

\noindent where $\delta_{i}$ is the mean fooling confidence increase of the post-softmax value of the target output neuron under the patch insertions for the $i^{th}$ synthetic adversary. If any $\delta$ is negative, we replace it with zero, and if the denominator is zero, we set $h_i$ to zero. 

Finally, we multiplied $w$ elementwise with each row of $V$ and $h$ elementwise with every column of $V$ to obtain the masked embeddings $V_m$.

\medskip

\subsection{Screening} \label{app:syn_nat}

\begin{figure*}[t!]
    \centering
    \includegraphics[width=\linewidth]{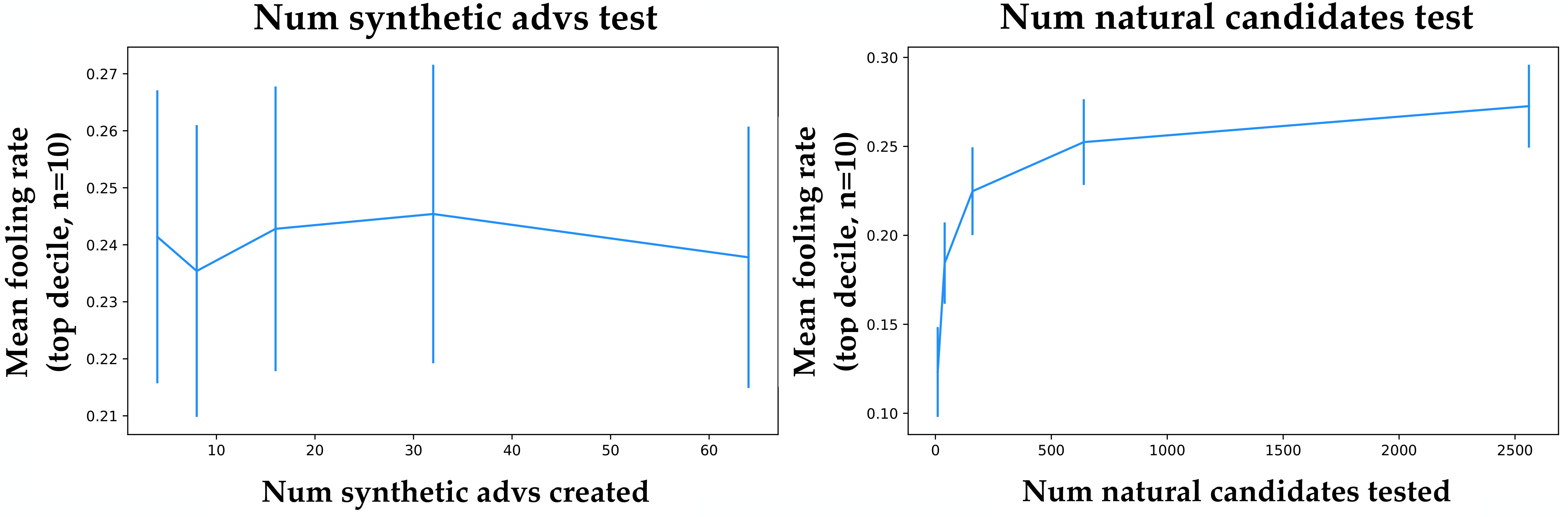}
    \caption{(Left) Mean natural patch success rate as a function of the number of synthetic adversaries we created, from which we selected the best 10 (or took all if there were fewer than 10) to then use in the search for natural patches. (Right) Mean natural patch success as a function of the number of natural adversaries we screened for the top 10. Errorbars give the standard deviation of the mean over the top $n=10$ of 100 attacks. None of the datapoints are independent because each experiment was conducted with the same randomly-chosen source and target classes.}
    \label{fig:syn_nat}
\end{figure*}

By default, for all experiments in this paper, we train 30 synthetic adversarial patches, select the most adversarial 10, then screen over 300 natural patches, and select the most adversarial 10.
These numbers were arbitrary, and because it is fully-automated, SNAFUE allows for flexibility in how many synthetic adversaries to create and how many natural adversaries to screen.
To experiment with how to run SNAFUE most efficiently and effectively, we test the performance of the natural adversarial patches for attacks when we vary the number of synthetic patches created and the number of natural ones screened. 
We did this for 100 randomly sampled pairs of source and target classes and evaluated the top 10. 
\figref{fig:syn_nat} shows the results.

As expected, when the number of natural patches that are screened increases, the performance of the selected ones increases. 
However, we find that creating more synthetic patches does not strongly influence the performance of the final natural ones. 
All of our choices of numbers of synthetic patches from 4 to 64 performed comparably well, likely due to redundancy.
One positive implication of this is that SNAFUE can be done efficiently with few synthetic adversaries. 
These were the main bottleneck in our runtime, so this has useful implications for speeding up runtimes. 
However, a negative implication of this is that redundant synthetic adversaries may fail to identify all possible weaknesses between a source and target class. 
Future work experimenting with the synthesis of \emph{diverse} adversarial patches will likely be valuable. 

\subsection{Are humans needed at all?} \label{app:humans}
SNAFUE has the advantage of not requiring a human in the loop -- only a human \emph{after} the loop to make a final interpretation of a set of images that are usually visually coherent. 
But can this step be automated too? 
To test this, we provide a proof of concept in which we use BLIP \cite{li2022blip} and ChatGPT \cite{schulman2022chatgpt} to caption the sets of images from the attacks in \figref{fig:nat_examples1}. 

First, we caption a set of 10 natural patches with BLIP \cite{li2022blip}, and second, we give them to ChatGPT (v3.5) following the prompt ``The following is a set of captions for images. Please read these captions and provide a simple "summary" caption which describes what thing that all (or most) of the images have in common.''

\begin{figure}[t!]
    \centering
    \includegraphics[width=0.9\linewidth]{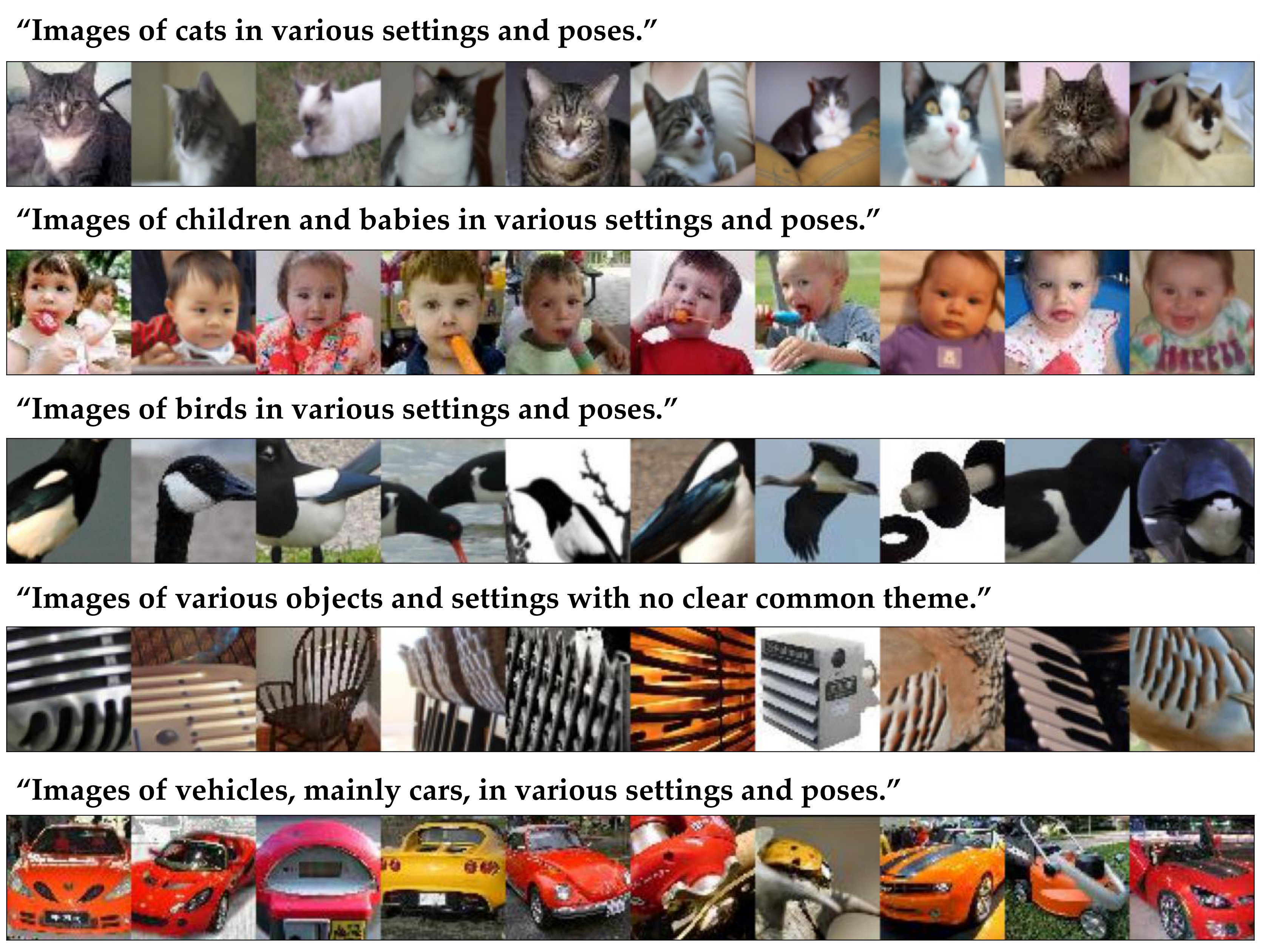}
    \caption{Natural adversarial patches from \figref{fig:nat_examples1} captioned with BLIP and ChatGPT.}
    \label{fig:captioning}
\end{figure}

Results are shown with the images in \figref{fig:captioning}.
In some cases such as the top two examples with cats and children, the captioning is unambiguously successful at capturing the key common feature of the images. 
In other cases such as with the black and white objects or the red cars, the captioning is mostly unsuccessful, identifying the objects but not the all of the key qualities about them.
Notably, in the case of the images with stripe/bar features, ChatGPT honestly reports that it finds no common theme. 
Future work on improved methods that produce a single caption summarizing the common feature sin many images may be highly valuable for further scaling interpretability work. 
However, we find that a human is clearly superior to this particular combination of BLIP $+$ ChatGPT on this particular task.

\subsection{Failure Modes for SNAFUE} \label{app:failures}

\begin{figure}[t!]
    \centering
    \includegraphics[width=0.9\linewidth]{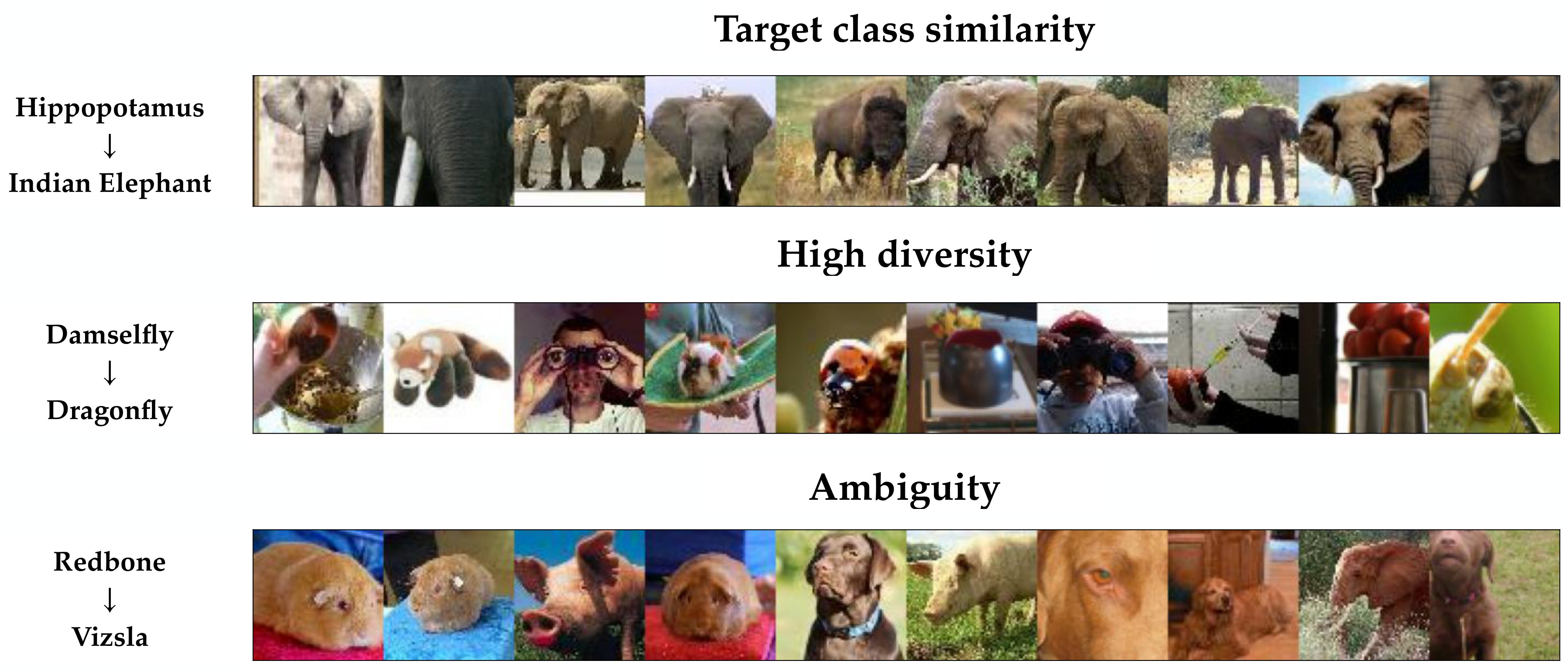}
    \caption{Examples of 3 of the 5 types of failure modes for SNAFUE that we describe in \secref{app:failures}.}
    \label{fig:failures}
\end{figure}

Here we discuss various non-mutually exclusive ways in which SNAFUE can fail to find informative, interpretable attacks.

\begin{enumerate}
    \item \textbf{An insufficient dataset:} SNAFUE is limited in its ability to identify bugs by the features inside of the candidate dataset. If the dataset does not have a feature, SNAFUE simply cannot find it. 
    \item \textbf{Failing to find adversarial features in the dataset:} SNAFUE will not necessarily recover an adversarial feature even if it is in the dataset.
    \item \textbf{Target class features:} Instead of finding novel fooling features, SNAFUE sometimes identifies features that simply resemble the target class yet evade filtering. \figref{fig:failures} (top) gives an example of this in which hippopotamuses are made to look like Indian elephants via the insertion of patches that evade filtering because they depict African elephants. 
    \item \textbf{High diversity:} We find some cases in which the natural images found by SNAFUE lack visual similarity and do not seem to lend themselves to a simple interpretation. One example of this is the set of images for damselfly to dragonfly attacks in \figref{fig:failures} (middle).
    \item \textbf{Ambiguity:} Finally, we also find cases in which SNAFUE returns a coherent set of natural patches, but it remains unclear what about them is key to the attack. \figref{fig:failures} (bottom) shows images for a redbone to vizsla attack, and it seems unclear from inspection alone the role that brown animals, eyes, noses, blue backgrounds, and green grass have in the attack because multiple images share each of these qualities in common.
\end{enumerate}

\subsection{Examples of Natural Adversarial Patches} \label{app:examples}

\begin{figure*}
    \centering
    \includegraphics[width=0.73\linewidth]{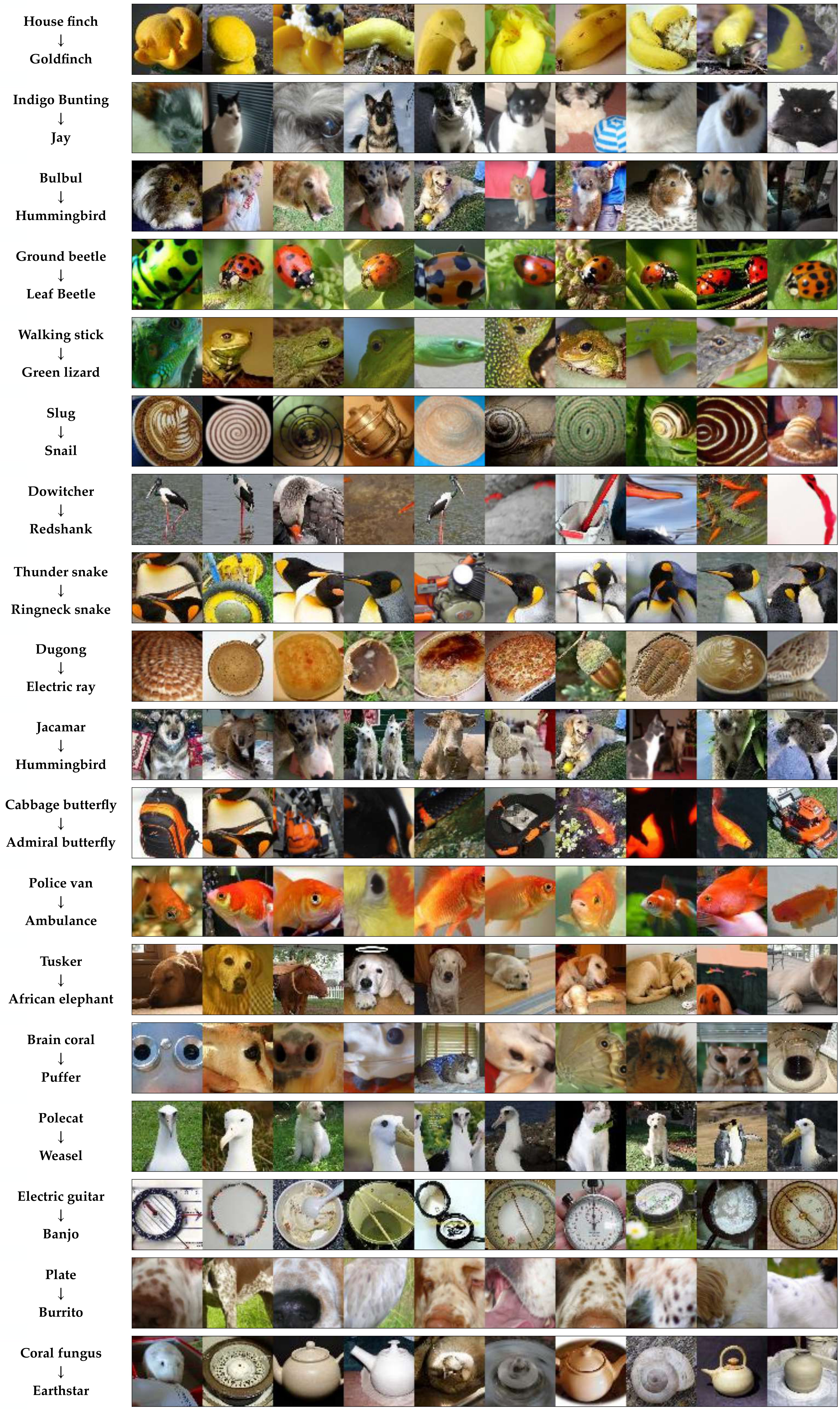}
    \caption{Examples of natural adversarial patches for several targeted attacks. Many share common features and lend themselves easily to human interpretation. Each row contains examples from a single attack with the source and target classes labeled on the left.}
    \label{fig:nat_examples2}
\end{figure*}

See \figref{fig:nat_examples2}.

\end{document}